\documentclass{article}
\usepackage{ijcai22}

\usepackage{amsmath,amsfonts,bm}

\def\Figref#1{Figure~\ref{#1}}

\def\eqref#1{equation~\ref{#1}}
\def\Eqref#1{Equation~\ref{#1}}

\def\Algref#1{Algorithm~\ref{#1}}

\def\1{\bm{1}}

\def\vtheta{{\bm{\theta}}}

\def\vg{{\bm{g}}}

\def\vx{{\bm{x}}}
\def\vy{{\bm{y}}}

\DeclareMathAlphabet{\mathsfit}{\encodingdefault}{\sfdefault}{m}{sl}
\SetMathAlphabet{\mathsfit}{bold}{\encodingdefault}{\sfdefault}{bx}{n}

\def\gB{{\mathcal{B}}}

\def\gS{{\mathcal{S}}}

\def\gX{{\mathcal{X}}}
\def\gY{{\mathcal{Y}}}

\newcommand{\E}{\mathbb{E}}

\newcommand{\KL}{D_{\mathrm{KL}}}

\usepackage{times}
\usepackage{soul}
\usepackage{url}
\usepackage[hidelinks]{hyperref}
\usepackage[utf8]{inputenc}
\usepackage[small]{caption}
\usepackage{graphicx}
\usepackage{amsmath}
\usepackage{amsthm}
\usepackage{booktabs}
\usepackage{algorithm}
\usepackage{algpseudocode}
\usepackage{mathrsfs}
\usepackage{multirow}

\newtheorem{definition}{Definition}
\newtheorem{lemma}{Lemma}

\title{Neighborhood Region Smoothing Regularization for Finding \\ Flat Minima In Deep Neural Networks}

\author{
Yang Zhao$^1$
\and
Hao Zhang$^1$
\affiliations
$^1$Department of Electronic Engineering, Tsinghua University\\
\emails
zhao-yan18@mails.tsinghua.edu.cn, \ haozhang@tsinghua.edu.cn
}

\begin{document}

\maketitle

\begin{abstract}
    Due to diverse architectures in deep neural networks (DNNs) with severe overparameterization, regularization techniques are critical for finding optimal solutions in the huge hypothesis space. In this paper, we propose an effective regularization technique, called Neighborhood Region Smoothing (NRS). NRS leverages the finding that models would benefit from converging to flat minima, and tries to regularize the neighborhood region in weight space to yield approximate outputs. Specifically, gap between outputs of models in the neighborhood region is gauged by a defined metric based on Kullback-Leibler divergence. This metric provides similar insights with the minimum description length principle on interpreting flat minima. By minimizing both this divergence and empirical loss, NRS could explicitly drive the optimizer towards converging to flat minima. We confirm the effectiveness of NRS by performing image classification tasks across a wide range of model architectures on commonly-used datasets such as CIFAR and ImageNet, where generalization ability could be universally improved. Also, we empirically show that the minima found by NRS would have relatively smaller Hessian eigenvalues compared to the conventional method, which is considered as the evidence of flat minima.

\end{abstract}

\section{Introduction}

Driven by the rapid development of computation hardwares, the scale of deep neural networks (DNNs) is unstoppably increasing, where the amount of parameters has significantly exceeded the sample size by even thousands of times  \cite{DBLP:conf/cvpr/HeZRS16,DBLP:conf/cvpr/HanKK17,DBLP:conf/iclr/DosovitskiyB0WZ21}. These heavily overparameterized DNNs would stay in huge hypothesis weight spaces. On the one hand, this provides the power to fit very complex or even arbitrary functions \cite{DBLP:journals/nn/HornikSW89}, on the other hand, it is challenging to seek optimal minima while resisting overfitting during training \cite{DBLP:conf/iclr/ZhangBHRV17}. In general, minimizing only the empirical training loss yielded by the difference between true labels and predicted labels (such as the categorical cross-entropy loss) could not provide sufficient guarantee on acquiring minima with satisfactory generalization ability. Regularization techniques are in higher demand than ever for guiding the optimizers towards finding minima with better generalization ability.

Based on the research, minima of well-generalized models are considered to be located at the loss landscape where the geometric characteristic of its neighborhood region tend to be flat \cite{DBLP:journals/neco/HochreiterS97a,DBLP:conf/iclr/KeskarMNST17}. However, it is nontrivial to explicitly finding flat minima in practice since the flatness of minima is demonstrated to be associated with the Hessian matrix of the loss function \cite{DBLP:conf/icml/DinhPBB17,DBLP:conf/iclr/KeskarMNST17}, and it is nearly impossible to solve such a huge Hessian matrix with current hardwares \cite{DBLP:conf/icml/BotevRB17}. Moreover, \cite{DBLP:conf/iclr/ForetKMN21,DBLP:conf/cvpr/ZhengZM21} empirically suggest that optimization via random perturbation within the neighborhood region in the weight space, although very simple to implement, would not help to flatten the loss landscape.

In this paper, we show that by adding proper regularization, such simple random perturbation could also lead to flat minima. Based on that, we propose a simple and effective regularization technique, called neighborhood region smoothing (NRS). NRS could regularize the models with random perturbation in the neighborhood region to yield the same outputs instead of their loss values. To accomplish such regularization, we firstly define the model divergence, a metric that gauges the divergence between models in the same weight space. We demonstrate that model divergence could provide interpretation of flat minima from the perspective of information theory, which shares similar core with the minimum description length principle. Since model divergence is differentiable, NRS regularization would enjoy minimizing the model divergence between models in the neighborhood region during optimizing for smoothing the surface around the target minima.

In our experiments, we show that NRS regularization could improve the generalization ability for various models on different datasets. Also, we confirm that for models with random perturbation in the neighborhood region, optimizing only the empirical loss could not lead to flat minima, just as \cite{DBLP:conf/iclr/ForetKMN21,DBLP:conf/cvpr/ZhengZM21} suggest. We empirically show that NRS regularization could reduce the largest eigenvalue of Hessian matrices, indicating that NRS regularization could indeed lead to flat minima.

\section{Related Works}

Training to make models generalize better is a fundamental topic in deep learning, where regularization is one of the most critical techniques that are commonly-used during training.

Regularization is actually a rather broad concept involving techniques that may be beneficial to the training process in various ways. Generally, common regularization techniques would affect the model training from three aspects. Firstly, some regularization expect to reduce the searching hypothesis space, where the most conventional method is weight decay \cite{DBLP:conf/nips/KroghH91,DBLP:conf/iclr/LoshchilovH19}. Secondly, some others try to complicate the target task so that the models could learn more "sufficiently". Typical methods include enlarging the input space such as cutout \cite{DBLP:journals/corr/abs-1708-04552} and auto-augmentation \cite{DBLP:journals/corr/abs-1805-09501}, and interfering models during training like dropout \cite{DBLP:journals/jmlr/SrivastavaHKSS14}, spatialdropout \cite{DBLP:conf/cvpr/TompsonGJLB15} regularization and stochastic depth regularization \cite{DBLP:conf/eccv/HuangSLSW16}. Thirdly, others implement normalizations, which includes batch normalization \cite{DBLP:conf/icml/IoffeS15}, layer normalization \cite{DBLP:journals/corr/BaKH16} and so on.

On the other hand, for a better understanding of generalization, researchers are also interested in studying the underlying factors that associate with the generalization ability of models. Several characteristics are demonstrated having impact on generalization including the flatness of minima, margins of classifiers \cite{DBLP:journals/tsp/SokolicGSR17,DBLP:conf/nips/BartlettFT17} and sparsity \cite{DBLP:conf/ijcai/Liu20a,zhao2021analyzing}. In particular, regarding the flatness of minima, despite still lack of theoretical justification and strict definition \cite{DBLP:conf/nips/NeyshaburBMS17}, empirical evidences have been found for such phenomenon. For example, \cite{DBLP:conf/iclr/KeskarMNST17} demonstrate that the reason large-batch training would lead to worse generalization than small-batch training is because large-batch training tends to stall at sharp minima. Moreover, by solving a minimax problem, \cite{DBLP:conf/iclr/ForetKMN21,DBLP:conf/cvpr/ZhengZM21,DBLP:journals/corr/abs-2110-03141} show that training could benefit from optimizing towards flat minima.

\section{Neighborhood Region Smoothing (NRS) Regularization}

Consider inputs $\vx \in \mathcal{X}$ and labels $\vy \in \mathcal{Y}$ which conform the distribution $\mathscr{D}$ and a neural network model $f$ parameterized by parameters $\vtheta$ in weight space $\mathbf{\Theta}$ which maps the given input space $\mathcal{X}$ to the corresponding output space $\mathcal{\hat{Y}}$,
\begin{equation}
    f(\cdot; \vtheta): \mathcal{X} \to \mathcal{\hat{Y}}
\end{equation}
\noindent For classification tasks, $\hat{\vy} \in \hat{\gY}$ could be a vector representing the predicted probability distribution of each class. 

Theoretically, we expect to minimize the expected loss $L_e(\vtheta) = \E_{\vx, \vy \sim  \mathscr{D}}[l(\hat{\vy}, \vy, \vtheta)]$. However, it is untractable in practice. We instead seek to acquire the model via minimizing the empirical loss $L_{\gS}(\vtheta) = \frac{1}{N} \sum_{i = 1}^{N} l(\hat{\vy}_i, \vy_i, \vtheta)$, where the training set $\gS = \{(\vx_{i}, \vy_{i})\}_{i = 0}^{N}$ is drawn independently and identically from distribution $\mathscr{D}$. In this way, gap between the expected loss and the empirical loss would directly lead to generalization errors of models.

\paragraph{Flat minima}
For over-parametrized models, the huge hypothesis weight space is filled with numerous minima. These minima may have approximate empirical training loss but diverse generalization ability in the meantime. In particular, minima of models with better generalization are supposed to locate at flatter surfaces, and these minima are often called flat minima. Although there may be different definitions for describing the flat minima \cite{DBLP:conf/iclr/KeskarMNST17,DBLP:conf/icml/DinhPBB17}, yet according to \cite{DBLP:journals/neco/HochreiterS97a}, the core interpretation behind conveys the similar idea that "a flat minimum is a large connected region in weight space where the error remains approximately constant". In other words, we expect that models parametrized by parameters in the neighborhood of $\vtheta$ could yield approximate outputs at flat minima. 

\paragraph{Model divergence}

Apparently, from previous demonstration, evaluating the flatness of minima would involve the assessment of how approximate the outputs are between models when given the same inputs. Simply using the loss gap between models as this assessment would be inappropriate because it is highly possible for two distinct outputs yield approximate loss value. To this end, we measure this by employing the Kullback-Leibler divergence,


\begin{definition}
    For model $f(\cdot; \vtheta)$ and model $f(\cdot; \vtheta^{'})$ in the same weight space $\mathbf{\Theta}$, given $\vx \in \gX$, the gap of the two models in the output space could be gauged via the Kullback-Leibler divergence,
    \begin{equation}
        d_p(\vtheta, \vtheta^{'}) = \E_{\vx}[\KL(f(\vx; \vtheta)||f(\vx; \vtheta^{'}))]
    \end{equation}
    \noindent where $\KL(\cdot || \cdot)$ denotes the KL divergence. We call $d_p(\vtheta, \vtheta^{'})$ the model divergence between $\vtheta$ and $\vtheta^{'}$.
\end{definition}

Note that $d_p(\vtheta, \vtheta^{'}) \geq 0$. In \cite{DBLP:conf/icml/DinhPBB17}, two models $f(\cdot; \vtheta)$ and $f(\cdot; \vtheta^{'})$ are said observationally equivalent if $f(\vx; \vtheta) = f(\vx; \vtheta)$ for $\forall \ \vx \in \gX$. Regarding model divergence, 

\begin{lemma}
    The two models $f(\cdot; \vtheta)$ and $f(\cdot; \vtheta^{'})$ are observationally equivalent if and only if $d_p(\vtheta, \vtheta^{'}) = 0$. 
\end{lemma}

Clearly, for $d_p(\vtheta, \vtheta^{'})$, the lower the value is, the more approximate the outputs that the two models could yield.

From the perspective of information theory, KL divergence of the two output distributions $f(\vx; \vtheta)$ and $f(\vx; \vtheta^{'})$ could be interpreted as how many additional bits are required to approximate the true distribution $f(\vx; \vtheta)$ when using $f(\vx; \vtheta^{'})$. Lower divergence indicates fewer information loss if using $f(\vx; \vtheta^{'})$ to approximate $f(\vx; \vtheta)$. 

So for flat minima, since the model is expected to have approximate outputs with models in its neighborhood region, they should have low model divergence. Based on the previous demonstration, when using models in the neighborhood to appropriate the true model, only few extra information is required. In contrast, high model divergence would mean more extra information is required. Therefore, describing a flat minimum would require much fewer information than a sharp minimum. Actually, this interpretation of flat minima from model divergence is in accordance with that in \cite{DBLP:journals/neco/HochreiterS97a}, where the authors suggest that flat minima should use few bits to describe from the perspective of minimum description length principle.

\paragraph{NRS regularization}
Generally, a flat minimum implies that for $\forall \ \delta \vtheta \in B(0, \epsilon)$ where $B(0, \epsilon)$ is the Euclidean ball centered at $0$ with radius $\epsilon$, the model $f(\cdot, \vtheta)$ and its neighborhood model $f(\cdot, \vtheta + \delta \vtheta)$ are expected to have both low model divergence and low empirical training loss. 

Therefore, we would manually add additional regularization in the loss during training for finding flat solutions,
\begin{equation}
    \label{eqn : final loss}
    \min_{\vtheta} L_{\gS}(\vtheta) + \alpha \cdot d_p(\vtheta, \vtheta + \delta \vtheta) + L_{\gS}(\vtheta + \delta \vtheta)
\end{equation}
\noindent where $\alpha$ denotes the penalty coefficient of model divergence $d_p(\vtheta, \vtheta + \delta \vtheta)$. In \Eqref{eqn : final loss}, the final training loss $L(\vtheta)$ contains three items: 
\begin{itemize}
    \item   The first item is the conventional empirical training loss, denoting the gap between the true labels and the predicted labels. Optimize this item would drive the predicted labels towards the true labels.
    \item The second item is the model divergence regularization, denoting the divergence between outputs yielded separately from the the models with parameter $\vtheta$ and the model with random perturbation $\delta \vtheta$ in the neighborhood of $\vtheta$. Optimize this item would explicitly drive the model to yield approximate outputs in the neighborhood.
    \item The third item is the empirical training loss of the neighborhood model, denoting the gap between true labels and the labels predicted by this neighborhood model. Optimize this item would explicitly force the neighborhood also learn to predict the true labels.
\end{itemize}

\paragraph{Practical implementation of NRS}
In order to provide sufficient $\delta \vtheta$ samples in neighborhood $B(0, \epsilon)$, it is best to train each input sample with a distinct $\delta \vtheta$. However, this would not suit the general mini-batch parallel training paradigm well in practice because this would require extra computation graphs for each distinct $\delta \vtheta$ when deploying using common deep learning framework like Tensorflow, Jax and Pytorch. For balancing the computation accuracy and efficiency, we would generate a unique $\delta \vtheta$ for each training device instead of for each input sample. In this way, each device need to generate only one neighborhood model, and all the input samples would fully enjoy the parallel computing in each device, which would significantly decrease the computation budget.


\Algref{alg:algorithm} shows the pseudo-code of the full implementation of NRS regularization. In \Algref{alg:algorithm}, the default optimizer is stochastic gradient descent. Besides, it should be also noted that we would normalize the $\delta \vtheta$ using $\vtheta$ at step 8 as we empirical find this would be helpful for training.

\begin{algorithm}[httb]
\caption{Neighborhood Region Smoothing (NRS) Regularization}
\label{alg:algorithm}
\textbf{Input}: Training set $\gS = \{(\vx_i, \vy_i)\}_{i = 0}^{N}$; loss function $l(\cdot)$; batch size $B$; learning rate $\eta$; total steps $K$; neighborhood region size $\epsilon$; model divergence penalty coefficient $\alpha$. \\
\textbf{Parameter}: Model parameters $\vtheta$. \\
\textbf{Output}: Model with final optimized weight $\hat{\vtheta}$.
\begin{algorithmic}[1] 
\State Parameter initialization $\vtheta_{0}$; get the number of devices $M$.
\For{step $k = 1$ to $K$}
\State Get sample batch $\gB = \{(\vx_i, \vy_i)\}_{i = 0}^{B}$.
\State Shard $\gB$ based on the number of devices $\gB = \gB_0 \cup \gB_1 \cdots \cup \gB_M$, where $| \gB_0 | = \cdots = |\gB_M|$.
\State \textbf{Do in parallel across devices.}
\State Make a unique pseudo-random number generator $\kappa$.
\State Generate random perturbation $\delta \vtheta$ within area $B(0, \epsilon)$ based on $\kappa$. 
\State Normalize $\delta \vtheta$ according to $||\vtheta_k||_2$, $\delta \vtheta = \frac{\delta \vtheta}{||\vtheta_k||_{2}}$.
\State Create neighborhood model $f(\cdot, \vtheta + \delta \vtheta)$. 
\State Compute gradient $\nabla_\vtheta L(\vtheta)$ of the final loss based on batch $\gB_i$.
\State \textbf{Synchronize and collect the gradient} $\vg$.
\State Update parameter $\vtheta_{k + 1} = \vtheta_{k} - \eta \cdot \vg$
\EndFor

\end{algorithmic}
\end{algorithm}

\begin{table*}[tb]
    \centering
    \begin{tabular}{lcccc}
    \toprule
        & \multicolumn{2}{|c|}{Cifar10} & \multicolumn{2}{|c|}{Cifar100} \\
    \midrule
    VGG16 & \multicolumn{1}{|c}{Basic} & \multicolumn{1}{c|}{Cutout} & \multicolumn{1}{|c}{Basic} & \multicolumn{1}{c|}{Cutout} \\
    \midrule
    Baseline & $\ 93.12_{ \pm 0.08}\ $ & $\ 93.95_{ \pm 0.11}\ $ & $\ 72.28_{ \pm 0.17}\ $  & $\ 73.34_{ \pm 0.22}\ $  \\
    RPR & $\ 93.14_{ \pm 0.21}\ $ & $\ 93.91_{ \pm 0.17}\ $ & $\ 72.31_{ \pm 0.21}\ $  & $\ 73.29_{ \pm 0.19}\ $  \\
    NRS & $\ \ \mathbf{93.79_{ \pm 0.11}}\ \ $ & $\ \  \mathbf{94.77_{ \pm 0.17}}\ \ $ & $\ \ \mathbf{73.61_{ \pm 0.20}}\ \ $  & $\ \ \mathbf{75.32_{ \pm 0.19}}\ \ $ \\
    \midrule
    ResNet18 & \multicolumn{1}{|c}{Basic} & \multicolumn{1}{c|}{Cutout} & \multicolumn{1}{|c}{Basic} & \multicolumn{1}{c|}{Cutout} \\
    \midrule
    Baseline & $\ 94.88_{ \pm 0.12}\ $ & $\ 95.45_{ \pm 0.16}\ $ & $\ 76.19_{ \pm 0.21}\ $  & $\ 77.03_{ \pm 0.11}\ $  \\
    RPR & $\ 94.90_{ \pm 0.24}\ $ & $\ 95.41_{ \pm 0.23}\ $ & $\ 76.25_{ \pm 0.31}\ $  & $\ 76.98_{ \pm 0.18}\ $  \\
    NRS & $\ \ \mathbf{96.01_{ \pm 0.15}}\ \ $ & $\ \  \mathbf{96.47_{ \pm 0.10}}\ \ $ & $\ \ \mathbf{78.67_{ \pm 0.17}}\ \ $  & $\ \ \mathbf{79.88_{ \pm 0.14}}\ \ $ \\
    \midrule
    WideResNet-28-10\ \ \ \  & \multicolumn{1}{|c}{Basic} & \multicolumn{1}{c|}{Cutout} & \multicolumn{1}{|c}{Basic} & \multicolumn{1}{c|}{Cutout} \\
    \midrule
    Baseline & $\ 96.17_{ \pm 0.12}\ $ & $\ 97.09_{ \pm 0.17}\ $ & $\ 80.91_{ \pm 0.13}\ $  & $\ 82.25_{ \pm 0.15}\ $  \\
    RPR & $\ 96.14_{ \pm 0.11}\ $ & $\ 97.02_{ \pm 0.15}\ $ & $\ 80.94_{ \pm 0.19}\ $  & $\ 82.19_{ \pm 0.24}\ $  \\
    NRS & $\ \ \mathbf{96.94_{ \pm 0.17}}\ \ $ & $\ \  \mathbf{97.55_{ \pm 0.13}}\ \ $ & $\ \ \mathbf{82.77_{ \pm 0.16}}\ \ $  & $\ \ \mathbf{83.94_{ \pm 0.16}}\ \ $ \\
    \midrule
    PyramidNet-164 & \multicolumn{1}{|c}{Basic} & \multicolumn{1}{c|}{Cutout} & \multicolumn{1}{|c}{Basic} & \multicolumn{1}{c|}{Cutout} \\
    \midrule
    Baseline & $\ 96.32_{ \pm 0.15}\ $ & $\ 97.11_{ \pm 0.12}\ $ & $\ 82.33_{ \pm 0.19}\ $  & $\ 83.50_{ \pm 0.17}\ $  \\
    RPR & $\ 96.31_{ \pm 0.19}\ $ & $\ 97.09_{ \pm 0.19}\ $ & $\ 82.25_{ \pm 0.24}\ $  & $\ 83.58_{ \pm 0.22}\ $  \\
    NRS & $\ \ \mathbf{97.23_{ \pm 0.19}}\ \ $ & $\ \  \mathbf{97.72_{ \pm 0.11}}\ \ $ & $\ \ \mathbf{84.61_{ \pm 0.24}}\ \ $  & $\ \ \mathbf{86.29_{ \pm 0.18}}\ \ $ \\
    \midrule
    ViT-B16 & \multicolumn{1}{|c}{Basic} & \multicolumn{1}{c|}{Heavy} & \multicolumn{1}{|c}{Basic} & \multicolumn{1}{c|}{Heavy} \\
    \midrule
    Baseline & $\ 80.04_{ \pm 0.11}\ $ & $\ 82.73_{ \pm 0.21}\ $ & $\ 54.42_{ \pm 0.25}\ $  & $\ 57.79_{ \pm 0.20}\ $  \\
    RPR & $\ 80.01_{ \pm 0.18}\ $ & $\ 82.73_{ \pm 0.17}\ $ & $\ 54.61_{ \pm 0.31}\ $  & $\ 57.81_{ \pm 0.17}\ $  \\
    NRS & $\ \ \mathbf{81.11_{ \pm 0.12}}\ \ $ & $\ \  \mathbf{84.03_{ \pm 0.16}}\ \ $ & $\ \ \mathbf{55.76_{ \pm 0.20}}\ \ $  & $\ \ \mathbf{59.52_{ \pm 0.22}}\ \ $ \\
    \bottomrule
    \end{tabular}
    \caption{Testing accuracy of various models on Cifar10 and Cifar100 when using the three training strategies.}
    \label{tbl : cifar}
\end{table*}

\section{Experimental Results}

In this section, we are going to demonstrate the effectiveness of NRS regularization by studying the image classification performances on commonly-used datasets with extensive model architectures. In our experiments, the datasets include Cifar10, Cifar100 and ImageNet, and the model architectures include VGG \cite{DBLP:journals/corr/SimonyanZ14a}, ResNet \cite{DBLP:conf/cvpr/HeZRS16}, WideResNet \cite{DBLP:conf/bmvc/ZagoruykoK16}, PyramidNet \cite{DBLP:conf/cvpr/HanKK17} and Vision Transformer \cite{DBLP:conf/iclr/DosovitskiyB0WZ21}. All the models are trained from scratch to convergence, which are implemented using Jax framework on the NVIDIA DGX Station A100 with four NVIDIA A100 GPUs.

\paragraph{Cifar10 and Cifar100}

We would start our investigation of NRS from evaluating its effect on the generalization ability of models on Cifar10 and Cifar100 dataset. Five network architectures would be trained from scratch, including VGG16, ResNet18, WideResNet-28-10, PyramidNet-164 and Vision Transformer B16\footnote{The names of all the mentioned model architectures represent the same meaning as them in their original papers.}. For datasets, we would adopt several different augmentation strategies. One is the \emph{Basic} strategy, which follows the conventional four-pixel extra padding, random cropping and horizontal random flipping, the other one is the \emph{Cutout} strategy, which would perform cutout regularization \cite{DBLP:journals/corr/abs-1708-04552} in addition to the basic strategy.

We would focus our investigations on the comparisons between three training strategies. The first one would train with the standard categorical cross-entropy loss. This is our baseline. The second one is to optimize the same cross-entropy loss of the models with random perturbation in the neighborhood region instead of the true model. This one is for confirming that such random perturbation could not be helpful to the generalization ability of models just as papers \cite{DBLP:conf/iclr/ForetKMN21,DBLP:conf/cvpr/ZhengZM21} suggest. We would call it RPR. The last strategy would use our NRS regularization. It should be noted that we would keep any other deployment the same for the three strategies during training except for the regularization mentioned in specific strategy.

For the common training hyperparameters, we would perform a grid search for acquiring best performance for each model except ViT-B16 model. Specifically, the base learning rate is searched over $\{0.01, 0.05, 0.1, 0.2\}$, the weight decay coefficient is searched over $\{0.0005, 0.001\}$ and the batch size is searched over $\{128, 256\}$. Also, we would adopt cosine learning rate schedule and SGD optimizer with 0.9 momentum during training. For ViT-B16 model, we use fixed hyperparameters where learning rate is $0.001$, weight decay is $0.3$, batch size is $256$ and the patch size is $4 \times 4$. Meanwhile, we would adopt the Adam optimizer during training. Additionally, for all the CNN and ViT models, we would use three different random seeds and report the average mean and variance across the testing accuracies of the three seeds. 

For RPR strategy, it involves one extra hyperparameter, the radius of the neighborhood region $\epsilon$. So similarly, we would perform a grid search over $\{0.05, 0.1, 0.5\}$ as well. As for NRS strategy, it involves two extra hyperparameters, the radius of the neighborhood region $\epsilon$ and the penalty coefficient of model divergence $\alpha$. We would adopt the same search for $\epsilon$, and $\alpha$ is searched over $\{0.5, 1.0, 2.0\}$. Notably, grid search of $\epsilon$ and $\alpha$ in both RPR and NRS would be performed on the basis of hyperparameters of the best model acquired by the first training strategy. 

Table \ref{tbl : cifar} shows the corresponding testing accuracies of Cifar10 and Cifar100, where all the reported results are the best results during the grid search of hyperparameters. We could see that in the table, all the testing accuracies have been improved by NRS regularization to some extent compared to the baseline, which confirms its benefit for model training. We also try NRS regularization on the recent Vision Transformer model. We could find that the testing accuracy of ViT model would be significantly lower than that of CNN models. This is because training ViT models generally requires plenty of input samples. Also, NRS could also improve the generalization ability of the such models. 

From the results, we could find that NRS regularization would not conflict with optimizers and current regularizations like dropout (in VGG16) and batch normalization, which is important for practical implementation. Additionally, we also confirm that simply optimizing the models with neighborhood random perturbation like RPR could indeed have no effect on improving the generalization ability of models.

\paragraph{ImageNet}

\begin{figure*}[t]
    \centering
    \includegraphics[width = 2\columnwidth]{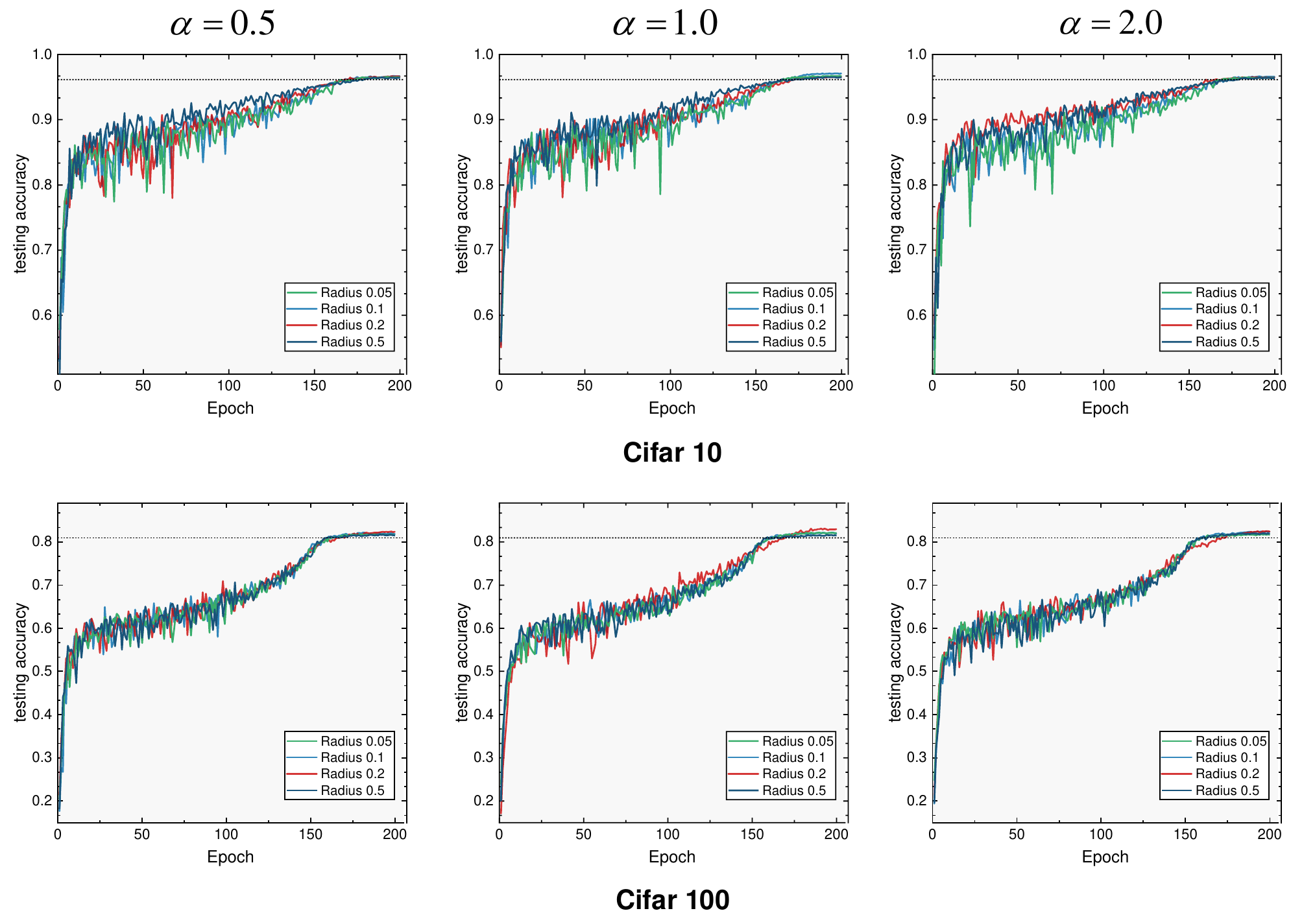}
    \caption{Evolutions of testing accuracy on Cifar10 and Cifar100 during training when trained with different parameters in NRS. The black dash lines refer to the reference lines which are the best testing accuracy trained using standard strategy.}
    \label{fig: study of parameters}
\end{figure*}

Next, we would check the effectiveness of NRS regularization on a large-scale dataset, ImageNet. Here, we would take VGG16, ResNet50 and ResNet101 model architectures as our experimental targets. For datasets, all the images would be resized and cropped to $224 \times 224$, and then randomly flipped in the horizontal direction. For common hyperparameters, instead of performing a grid search, we would fix the batch size to 512, the base learning rate to 0.2, the weight decay coefficient to 0.001. During training, we would smooth the label with 0.1. All models would be trained for a total of 100 epochs with three different seeds. 

Our investigation would focus on the comparisons between two training strategies. One is the standard training with categorical cross-entropy loss, which is our baseline. The other one is trained with NRS regularization. For hyperparameters in NRS regularization, we would fix the $\epsilon$ to 0.1 and $\epsilon$ to 1.0. Table \ref{tbl : imagenet} shows the corresponding results. 

\begin{table}[httb]
    \centering
    \begin{tabular}{lcc}
        \toprule
        & \multicolumn{2}{|c|}{ImageNet} \\
        \midrule
        VGG16 & \multicolumn{1}{|c}{Top-1 Accuray} & \multicolumn{1}{c|}{Top-5 Accuray} \\
        \midrule
        Baseline & $\ 73.11_{ \pm 0.08}\ $ & $\ 91.12_{ \pm 0.08}\ $   \\
        NRS & $\ \mathbf{73.52_{ \pm 0.09}}\ $ & $\ \mathbf{91.39_{ \pm 0.09}}\ $ \\
        \midrule
        ResNet50 \ \ \ & \multicolumn{1}{|c}{Top-1 Accuray} & \multicolumn{1}{c|}{Top-5 Accuray} \\
        \midrule
        Baseline & $\ 75.45_{ \pm 0.15}\ $ & $\ 93.04_{ \pm 0.07}\ $   \\
        NRS & $\ \mathbf{76.29_{ \pm 0.13}}\ $ & $\ \mathbf{93.53_{ \pm 0.10}}\ $ \\
        \midrule
        ResNet101 \ \ \ & \multicolumn{1}{|c}{Top-1 Accuray} & \multicolumn{1}{c|}{Top-5 Accuray} \\
        \midrule
        Baseline & $\ 77.15_{ \pm 0.11}\ $ & $\ 93.91_{ \pm 0.07}\ $   \\
        NRS & $\ \mathbf{78.02_{ \pm 0.10}}\ $ & $\ \mathbf{94.44_{ \pm 0.08}}\ $ \\
        \bottomrule
    \end{tabular}
    \caption{Testing accuracy of various models on ImageNet dataset when using standard training strategy (Baseline) and NRS regularization strategy.}
    \label{tbl : imagenet}
\end{table}

As we could see in Table \ref{tbl : imagenet}, the testing accuracy could be improved again to some extent when using NRS regularization. This further confirms that NRS regularization could be beneficial to the generalization ability of models.

\section{Further Studies of NRS}

\subsection{Parameter Selection in NRS}

In this section, we would investigate the influence of the two parameters $\epsilon$ and $\alpha$ in NRS regularization on the results. The investigation is conducted on Cifar10 and Cifar100 using WideResNet-28-10. We train the models from scratch using NSR with the same common hyperparameters of the best models acquired by the baseline strategy. And then, we would perform the grid search for the two parameters using the same scheme as in previous section.

\Figref{fig: study of parameters} shows the evolution of testing accuracy during training when using the corresponding different parameters. We could see that actually for all the deployed $\epsilon$ and $\alpha$, the generalization ability could be somewhat improved on both Cifar10 and Cifar100 datasets compared to the baseline (the black reference line in the figure). This again demonstrates the effectiveness of NRS regularization. From the figure, we could find that the model could achieve the best performance when $\epsilon = 0.1$ and $\alpha = 1.0$ for Cifar10 dataset and $\epsilon = 0.2$ and $\alpha = 1.0$ for Cifar100 dataset. Also, we could find that the generalization improvement on Cifar100 would generally be larger than that on Cifar10.

\subsection{Eigenvalues of Hessian Matrix}

In this section, we would investigate the intrinsic change of models when using the NSR regularization. In general, the eigenvalues of Hessian matrix are considered to have connections with the flatness of minima. Specifically, the largest eigenvalue of Hessian matrix of flat minima could be larger than that of sharp minima.

Since the dimension of the weight space is so huge, solving the Hessian matrix in a direct manner is nearly impossible. Therefore, we would employ the approximate method introduced in \cite{DBLP:conf/icml/BotevRB17}. It calculates the diagonal block of Hessian matrix recursively from the deep layers to the shallower layers.

We would still use WideResNet-28-10 model and Cifar10 as our investigation target. We would use the best models acquired by the three training strategies in the previous section. Also, we would calculate the eigenvalues of Hessian matrix for the last layer in each model. Table \ref{tbl : eigenvalues} reports the results.

\begin{table}[httb]
    \centering
    \begin{tabular}{lcc}
        \toprule
        WideResNet-28-10 & \multicolumn{1}{|c}{$\lambda_{max}$}  \\
        \midrule
        Baseline & \ \ \ \ \ \  49.15  \ \ \ \ \ \ \\
        RPR & \ \ \ \ \ \  51.79  \ \ \ \ \ \ \\
        NRS & \ \ \ \ \ \  \textbf{2.42}  \ \ \ \ \ \ \\
        \bottomrule
    \end{tabular}
    \caption{The largest eigenvalue of Hessian matrix of models trained with three different strategies.}
    \label{tbl : eigenvalues}
\end{table}

As can be seen from the table, using NRS regularization can significantly reduce the largest eigenvalue of the Hessian matrix compared to using the other two strategies. Also, using RPR strategy indeed could not lead to flat minima. This again verify that using NRS regularization would find flat minima during training.

\section{Conclusion}

In this paper, we propose a simple yet effective regularization technique, called Neighborhood Region Smoothing, for finding flat minima during training. The key idea of NRS is regularizing the neighborhood region of models to yield approximate outputs. Using outputs in NRS could give stronger regularization than using loss values, so simple random perturbation in the neighborhood region would be effective. We define model divergence to gauge the gap between outputs of models in the neighborhood region. In this way, NRS regularization is achieved by explicitly minimizing both the empirical loss and the model divergence. In our experiments, we show that using NRS regularization could improve the generalization ability of a wide range of models on diverse datasets compared to two other training strategies. We also investigate to give the best hyperparameters in NRS on Cifar10 and Cifar100 dataset. Finally, smaller eigenvalue of Hessian matrix confirms that NRS regularization could indeed to flat minima.

\bibliographystyle{named}
\bibliography{main}

\end{document}